\documentclass[10pt,twocolumn,letterpaper]{article}

\usepackage[final,algorithms]{wacv}

\usepackage[margin=0.7in]{geometry}
\usepackage[utf8]{inputenc}
\usepackage[T1]{fontenc}
\usepackage{amsmath,amssymb,amsfonts}
\usepackage{booktabs}
\usepackage{microtype}
\usepackage{nicefrac}
\usepackage{xcolor}
\usepackage{graphicx}
\usepackage{subcaption}
\usepackage{algorithm}
\usepackage{algorithmic}
\usepackage{hyperref}
\usepackage{url}

\hypersetup{
  colorlinks=true,
  linkcolor=blue,
  citecolor=blue,
  urlcolor=blue
}

\graphicspath{{./figures/}}

\title{Exploring Embedding Priors in Prompt-Tuning for Improved Interpretability and Control}

\author{
Sergey Sedov\\
New York University, NY, USA\\
\texttt{ss19021@nyu.edu}
\and
Venu Gopal Kadamba\\
New York University, NY, USA\\
\texttt{vk2636@nyu.edu}
\and
Sumanth Bharadwaj Hachalli Karanam\\
New York University, NY, USA\\
\texttt{sh8111@nyu.edu}
}

\date{}

\begin{document}
\maketitle

\begin{abstract}
Prompt-Tuning is an efficient method for adapting pre-trained language models to new tasks with minimal computational overhead by modifying prompt embeddings. In this work we investigate how crucial the embedding collapse, frequently observed in Prompt-Tuning, is for the final performance of the model. To answer this question, we design embedding priors and compare them with posteriors of the converged Soft and Deep Prompt-Tuning. Our findings suggest that priors strongly affect the position of the tuned embeddings, and models can effectively work with embeddings from different parts of activation spaces, including the completely new regions. As final Prompt-Tuning capabilities are limited, we hypothesize that controllable Prompt-Tuning posteriors may serve as a good starting point in such tasks as COT distillation. Our experiments also show that generated trajectories are not localized in activation space of the models. However, there are distinct clusters of activations for distant tasks (NLP and arithmetics), while activations between NLP tasks lie in the same cluster (Question-Answering and MLM). These observations leave us questioning the importance of single activation cluster for generalization abilities of large language models.
\end{abstract}

\section{Introduction}

Prompt-tuning has emerged as a powerful approach for adapting pre-trained language models to new tasks with minimal computational overhead. By modifying the model's prompt embeddings, this technique provides a flexible mechanism for task-specific adaptation. However, a critical issue with prompt-tuning is the phenomenon of embedding collapse, where the embeddings of newly tuned tokens tend to converge toward specific, pre-existing token embeddings. This clustering reduces the diversity of embeddings, limiting the model’s ability to generalize across different language domains and leading to overfitting on task-specific features.

The central question of this research is: To what extent can we control the distribution of prompt-tuned embeddings to avoid embedding collapse, and how does this impact the model's generalization capabilities? We propose leveraging Bayesian priors to influence the embedding space during prompt-tuning, guiding the model toward more flexible, interpretable, and adaptable embeddings. While the initial hypothesis of this research is that by mitigating embedding collapse, we could allow the model to better generalize across a variety of tasks and domains, we also aim to better interpret pre-trained models' behavior on activation spaces by experimenting with their priors.

Our research focuses on the following key objectives:
\begin{itemize}
\item
Examining pre-trained model activations distributions on datasets that differ to the ones used on pre-training.
\item
Investigating how different priors influence the learned embeddings during prompt-tuning.
\item
Exploring the divergence of prompt-tuned embeddings from pre-trained token embeddings and assessing the impact of such divergence on model performance.
\item 
Designing embeddings that support interpolation and generalization across multiple language domains.
\end{itemize}

\subsection*{Why do we need to control the posterior?}
Generally speaking, the findings of this research could help improve the interpretability and cross-domain adaptability of prompt-tuning methods, contributing to more robust and flexible language models. However, we further point out the possible impact of this research on other fields.

Following the hypothesis that the behavior of the Prompt-Tuning methods is more interpretable than other adapter methods, we propose to use Prompt-Tuning as the toy setup, to test the generalization abilities of the language models to work on different data. For instance, we think about the future applications of this research in such fields as Chains-Of-Thoughts distillation and expansion of language models to the Multi-Modality tasks. 

While research on Prompt-Tuning posteriors may bring some general evidence about the behavior of such embedding space expansion methods, we propose that controlled prompt-tuning posteriors may then be used as the prior distributions for the embeddings in these tasks.

Our key takeaways regarding activations distributions and prompt-tuning control may be summarized as follows:
\begin{itemize}
\item Sentence trajectories are generally not localized, neither on token embeddings nor on deep activation levels.
    \item Deep activations for distant tasks have distinct clusters.
    \item Prompt-Tuning shows the same quality on embeddings from different parts of activation space, including the completely new ones for the model.
    \item Priors strongly affect the position of the trained embeddings. However, we are not able to improve Prompt-Tuning results starting from different priors, which suggests that we are able to train embeddings to their full capabilities both in and out of their initial cluster.

\end{itemize}

\section{Related Work}

Prompt-tuning has indeed gained significant attention as an efficient method for adapting large pre-trained models to specific tasks. This approach offers several advantages over traditional fine-tuning, particularly in terms of computational efficiency and parameter efficiency. Recent research has expanded on the concept of prompt-tuning, introducing more sophisticated techniques to optimize the tuning process.

Deep Prompt Tuning (DPT), as proposed by Liu et al. (2021) \cite{liu2021prompt}, represents a significant advancement in this field. DPT extends the concept of prompt-tuning by inserting trainable prompt tokens at multiple layers of the model, rather than just at the input. This approach allows for more fine-grained control over the model's behavior and has shown impressive results across various natural language processing tasks.

The issue of embedding collapse in prompt-tuning, first noted by Lester et al. (2021) \cite{lester2021power}, remains a significant challenge. This phenomenon occurs when the newly tuned embeddings cluster too closely with pre-existing token embeddings, limiting their flexibility and generalization capabilities. The problem bears similarities to neural collapse observed in vision models, where class representations become highly specialized.

To address these challenges, researchers have begun exploring Bayesian methods in the context of prompt-tuning. For instance, Lee et al. (2024) introduced Bayesian Multi-Task Transfer Learning for Soft Prompt Tuning (BMTPT) \cite{lee2024bayesian}. This approach uses Bayesian principles to model the posterior distribution of prompts across multiple source tasks, using this distribution as a prior for target tasks. By employing Stein Variational Gradient Descent (SVGD), BMTPT approximates the source prompts' posterior distribution, potentially offering a more robust initialization for target tasks.

Another interesting development is the application of Bayesian principles to prompt learning in image-language models. Derakhshani et al. (2022) \cite{derakhshani2022bayesian} proposed a Bayesian prompt learning framework that models the input prompt space as a probabilistic distribution. This approach aims to regularize the prompt space, reduce overfitting to seen prompts, and improve generalization to unseen prompts. By framing prompt learning as a variational inference problem, they introduce a regularization term that encourages the model to learn informative prompts scattered across the prompt space.

These Bayesian approaches to prompt-tuning and prompt learning offer promising directions for addressing the challenges of embedding collapse and improving generalization. By incorporating prior knowledge and managing uncertainty in the embedding space, these methods have the potential to create more flexible and interpretable representations. As research in this area continues to evolve, we can expect to see further refinements and novel applications of Bayesian principles in the context of prompt-tuning and related techniques.

\section{Methodology}

In this project, we are exploring the use of Bayesian priors in Prompt-Tuning for Question-Answering and arithmetic tasks using the LLaMA 3.2 1B model, which has 16 layers. We are making use of Stanford Question Answering Dataset (SQuAD) and DeepMind MATH Dataset (arithmetic subtask) for the experiments. 

\subsection{Prompt-Tuning}
In our Prompt-Tuning experiments we are training 20 token embeddings, prepending them to the original model input, while all model weights are frozen.

\subsection{Deep Prompt-Tuning}
In our Deep Prompt-Tuning experiments we are training 20 activation-level embeddings on 3 last layers of the model and 20 token embeddings as well. Specifically, we modify model forward to prepend trainable embeddings to each layer input and cut the first 20 embeddings of layer output, so that generally model is processing only the initial sequence length on other layers. It differs from the original setup of Prompt-Tuning, as there the sequence length is extended with 20 embeddings for each layer of the model. However, our experiments show no difference between these setups in terms of model performance when only token embeddings are trained. 

\subsection{Prior Design}
We experiment with different prompt embedding priors:
\begin{itemize}
    \item \textbf{Gaussian Priors}: To serve as a baseline, we will use isotropic Gaussian priors. These will help us assess whether simple, unstructured priors lead to embedding collapse.

    The isotropic Gaussian prior is defined as:
    \[ p(\mathbf{e}) = \mathcal{N}(\mathbf{e}; \mathbf{0}, \sigma^2 \mathbf{I}) \]
    where \( \mathbf{e} \) is sampled embedding, and \(\sigma^2\) is scalar.

    \item \textbf{Structured Priors}: Inspired by the work of Zhu et al. (2020), we will explore structured priors that avoid placing embeddings too close to pre-existing clusters of token embeddings. These priors will encourage the model to sample embeddings from less dense regions in the embedding space.

    The structured Gaussian prior, which considers correlations between dimensions, is given by:
    \[ p(\mathbf{e}) = \mathcal{N}(\mathbf{e}; \mathbf{\mu}, \mathbf{\Sigma}) \]
    where \( \mathbf{\mu} \) is the mean vector, and \( \mathbf{\Sigma} \) is the covariance matrix. The parameters \( \mathbf{\mu} \) and \( \mathbf{\Sigma} \) can be estimated as:
    \[ \mathbf{\mu} = \frac{1}{N} \sum_{i=1}^N \mathbf{e}_i, \quad \mathbf{\Sigma} = \frac{1}{N} \sum_{i=1}^N (\mathbf{e}_i - \mathbf{\mu})(\mathbf{e}_i - \mathbf{\mu})^T \]
    where \( N \) is the number of pre-trained embeddings, and \( \mathbf{e}_i \) is the \( i \)-th embedding.

    Besides Gaussian prior, we experimented with Gaussian exclusion and Gaussian interpolation priors. These results are presented in Sections 4.3 and 4.5. 

    In Gaussian exclusion, we fit Gaussian $\mathcal{N}(\mu, \Sigma)$ on embeddings data, and sample from wider distribution: $x \sim \mathcal{N}(\mu, c_{\mathrm{dim}} \cdot \Sigma)$, where in our experiments
    \[ c_{\mathrm{dim}} = \exp \big((2 / \mathrm{dim}) \cdot \log(c)\big), \quad c = 5. \]
    After that, we accept samples with probability: $p = \max \big(0, 1 - \mathrm{PDF}(x) / \mathrm{PDF}_{\mathrm{wide}}(x) \big)$.

    In Gaussian interpolation, we propose to interpolate samples between Gaussians fitted on the new domain and the pre-training domain. Specifically, for the $i$-th pair of samples from two Gaussians $x_i \sim \mathcal{N}(\mu_1, \Sigma_1)$ and $y_i \sim \mathcal{N}(\mu_2, \Sigma_2)$ we take
    \[ e_i = \alpha \cdot x_i + (1 - \alpha) \cdot y_i, \quad \alpha \sim U[0, 1]. \]

    Moving from Gaussian approximation of activations distribution, we also experimented with VAE-sampled embeddings, hypothesizing that VAE can smoothen distributions between different domains. Results for this method are presented in Section 4.6.
\end{itemize}

\subsection{Do we really need to control the prior?}

The central hypothesis of our work is that embedding priors can guide the model toward more adaptable and generalizable embeddings, preventing collapse into task-specific clusters. To test this, we measure the divergence between the prompt-tuned embeddings and the pre-trained token embeddings. We use t-SNE visualizations to observe the clustering behavior of embeddings and quantify how far the prompt-tuned embeddings deviate from the original token embeddings. Following the side-motivation of the controllable interpretability of the adapter-like methods, our experiments show that prior designs influence the trained embeddings and final quality of the model. On the specific tasks, models sometimes have meaningfully clusterized activations, so it would be interesting to design such priors too.


\subsection{Experiment Setup}

We use the LLaMA-1B model to train a Question Answering Task using prompt-tuning. The task is evaluated on a Stanford Question Answering dataset (SQuAD), with performance measured in terms of accuracy, precision, recall, and F1 score. In our experiments we aim to vary the prior types (Gaussian vs. structured priors) and compare their effects on the performance and diversity of the embeddings. Further on, we investigate the influence of regularization techniques on the posterior conjunction.

\section{Experiments}

To validate the proposed methodology, we are conducting series of experiments.

\subsection{Proof of concept experiments}
\begin{figure}[ht]
        \centering
        \includegraphics[width=\linewidth]{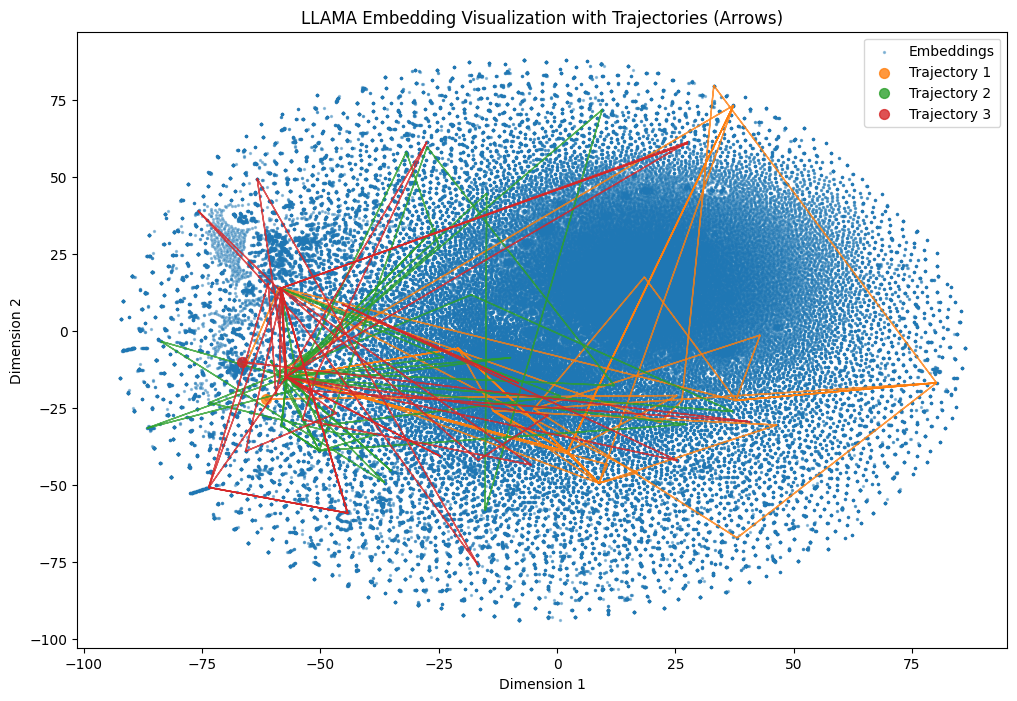}
        \caption{Sentences on the LLaMA token embedding space, t-SNE}
        \label{fig:token-walks}
    \end{figure}

In this section we aim to better understand the task complexity and which problems we may face while designing priors. 
\begin{itemize}
    \item \textbf{Plotting sentences on token embedding space.}
    
    We start by plotting the token trajectories on the t-SNE scatter-plot of pre-trained embeddings in Figure \ref{fig:token-walks}. In this section we are measuring how local the model steps on the token-embedding level and whether we need to condition on these trajectories in the designs of our priors for the specific tasks. We can see that trajectories are highly jumpy. Their standard deviations tend to be statistically lower than the random walks across vocabularies, though close enough to them. 
\end{itemize}

    \begin{figure}[ht]
        \centering
        \includegraphics[width=\linewidth]{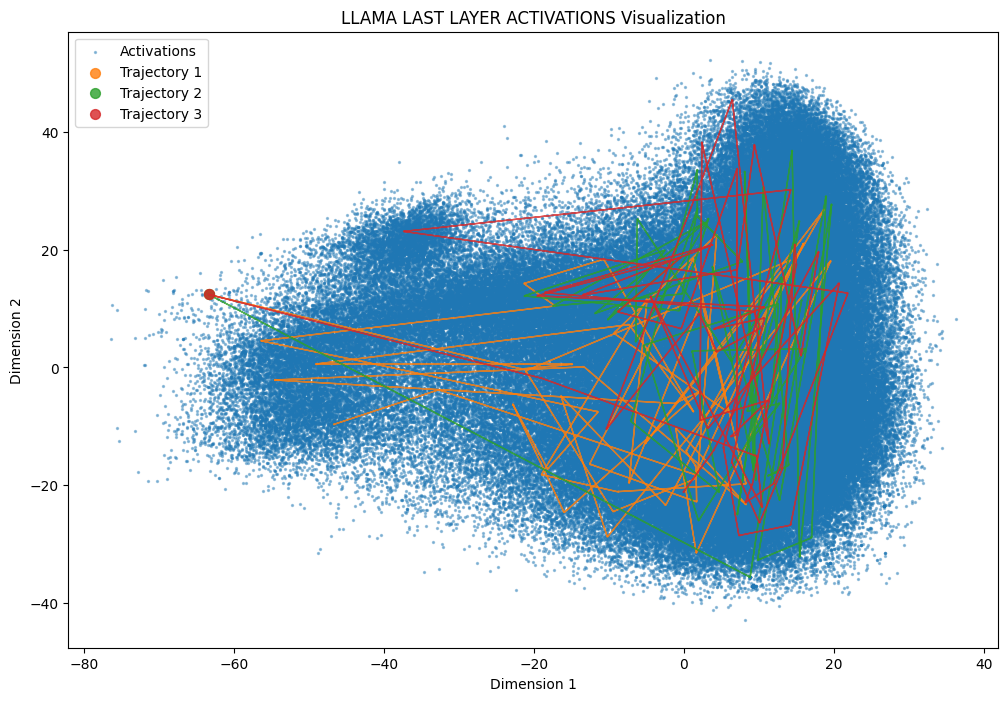}
        \caption{Sentences on the LLaMA activation space, PCA}
        \label{fig:activations-walks}
    \end{figure}

\begin{figure}[ht]
        \centering
        \includegraphics[width=\linewidth]{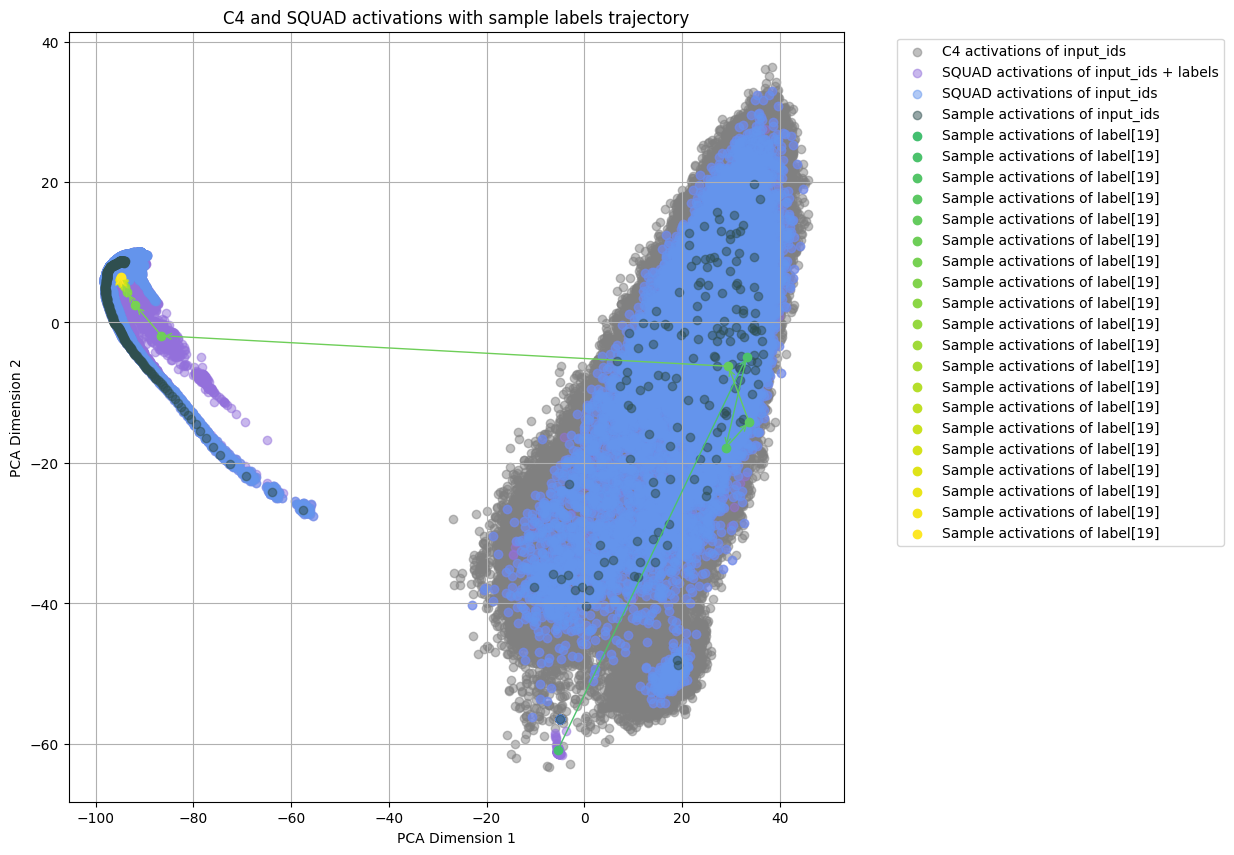}
        \caption{SQuAD answers trajectories on the LLaMA token last layer activation space. Gray is C4, blue is SQuAD, dark-green is sampled question, green lines represent next-token answer generation. Left tail corresponds to PAD tokens. PCA}
        \label{fig:activations-labels}
    \end{figure}

\begin{itemize}
    \item \textbf{Plotting sentences on activations space.}

    We proceed with making a similar plot in Figure \ref{fig:activations-walks} on the higher-layer activations of the model, gathered on the 1000 random samples from the C4 dataset. Trajectories still do not localize well, which means that while designing priors we should either look for the specific tasks, or focus on general features of the prior, e.g. intersection with the activations distributions.

    We further look at the specific Question-Answering task in SQuAD in Figure \ref{fig:activations-labels}, representing the last layer activations distribution. First of all, interestingly SQuAD activations distribution does not diverge far from C4. Secondly, we can observe that 4 meaningful tokens in the answer lie closely in the activation space, though their set tightly intersects with question activations.
\end{itemize}

\begin{figure}[ht]
        \centering
        \includegraphics[width=\linewidth]{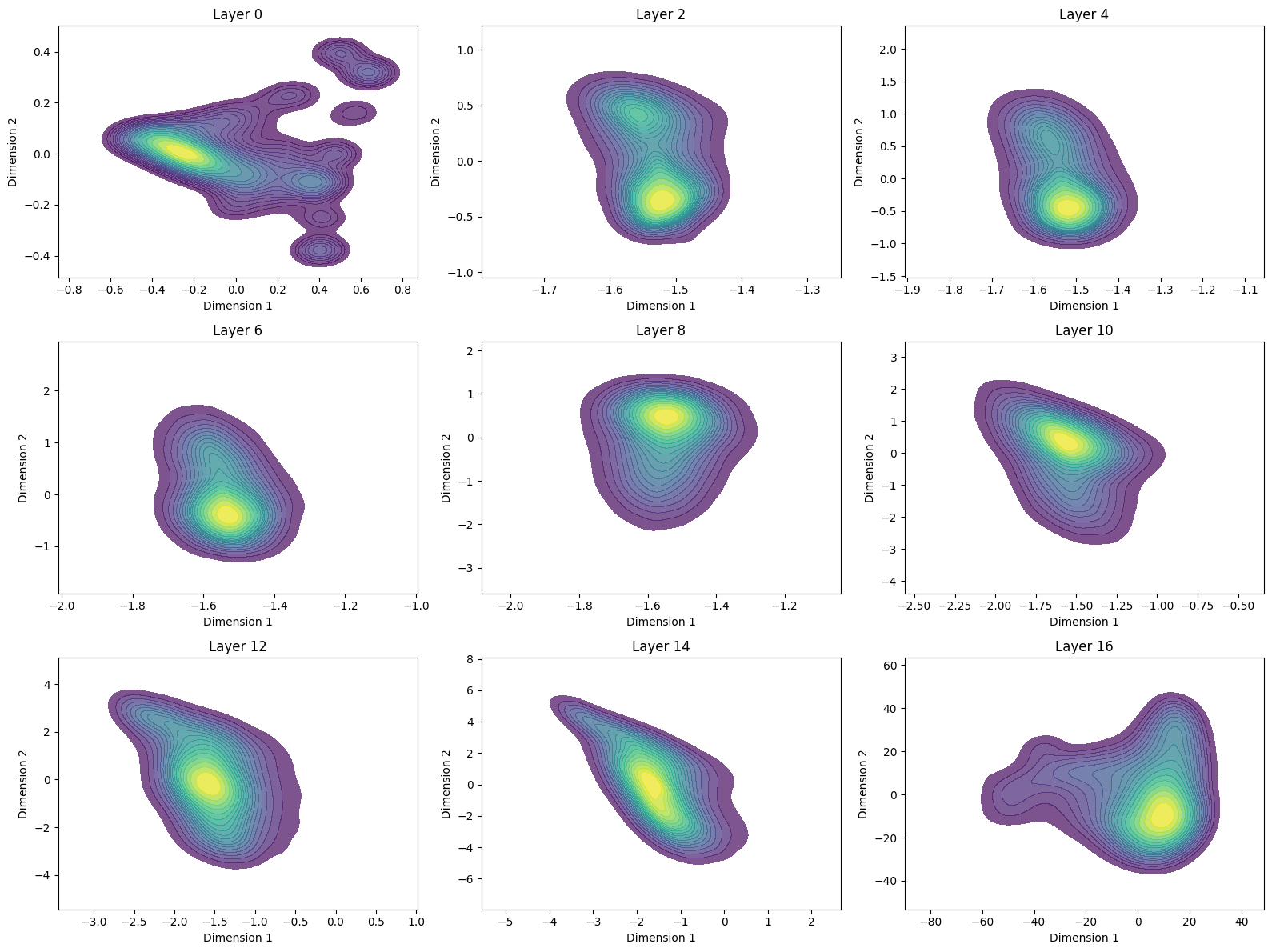}
        \caption{Distributions of LLaMA activation space, PCA}
        \label{fig:activations-layers}
    \end{figure}

\begin{itemize}
    \item \textbf{Different layers activations}
    
    We further check that the distribution of activations stays more or less unimodal across different layers in Figure \ref{fig:activations-layers}.
\end{itemize}

\begin{figure}[ht]
        \centering
        \includegraphics[width=\linewidth]{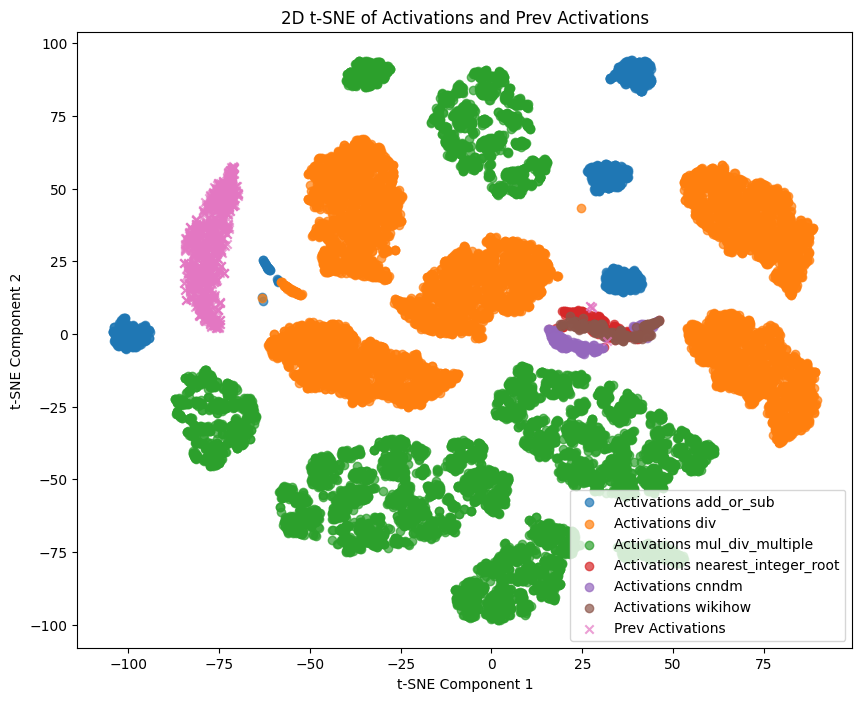}
        \caption{Mean activations of the T5-base on C4 (pink) and MATH datasets, t-SNE}
        \label{fig:t5-tsne}
    \end{figure}

\begin{itemize}
    \item \textbf{Different Modalities across language tasks}
    
    Finally, we include the t-SNE plot of the T5-base model mean activations in Figure \ref{fig:t5-tsne}, gathered over the 1000 samples of C4-dataset and 6 different arithmetics tasks from DeepMind MATH dataset. This experiment brings the evidence that the localized prior design depending on the specific tasks may be appropriate. In this case, mean activations over math problems are far away from the original C4 activations. Moreover, different math tasks are far apart between each other, which is probably the evidence of poor generalization abilities of this model across math tasks.
    
    Though we have not conducted this experiment on the LLaMA model and most likely modern models possess better abilities on math tasks, we provide this experiment as an illustration of the potential of prompt-tuning control. LLMs possess weak generalization abilities across different subsets of a relatively new domain of data, that could be enhanced by "bridging the gap" between these subsets with additional tokens. 
\end{itemize}

\subsection{Token-Embedding Priors and Learning Rate}
\begin{figure}[ht]
    \includegraphics[width=\linewidth]{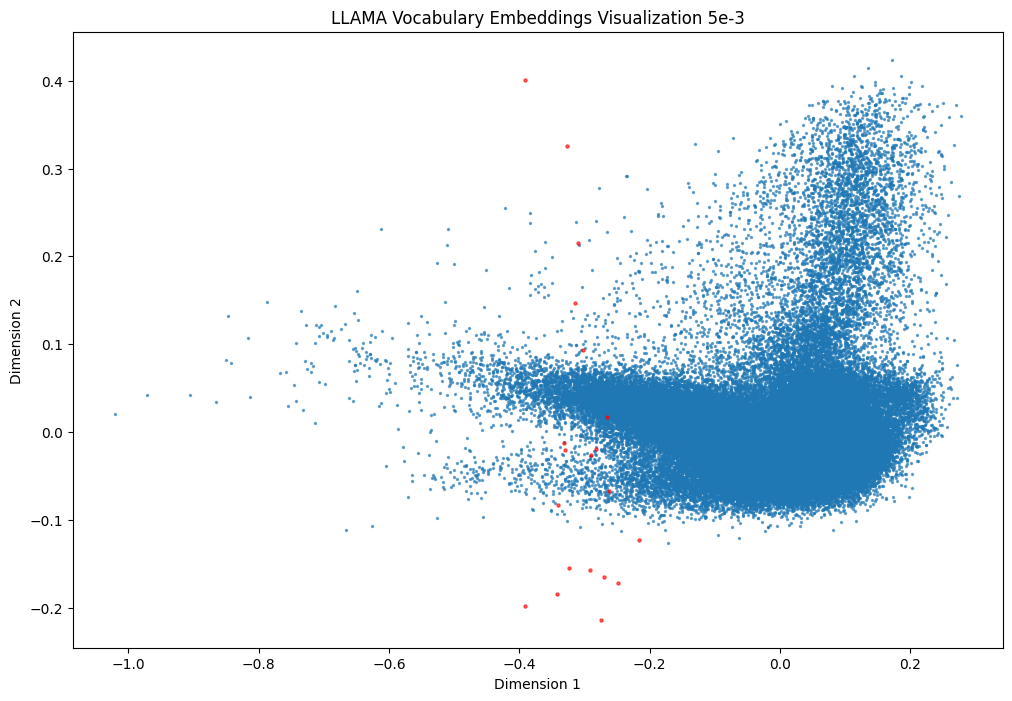}
    \caption{Isotropic Gaussian Prompt Initialization with 5e-3 learning rate (all token embeddings)}
    \label{fig:isotropic-gaussian-5em3}
\end{figure}

\begin{figure}[ht]
    \includegraphics[width=\linewidth]{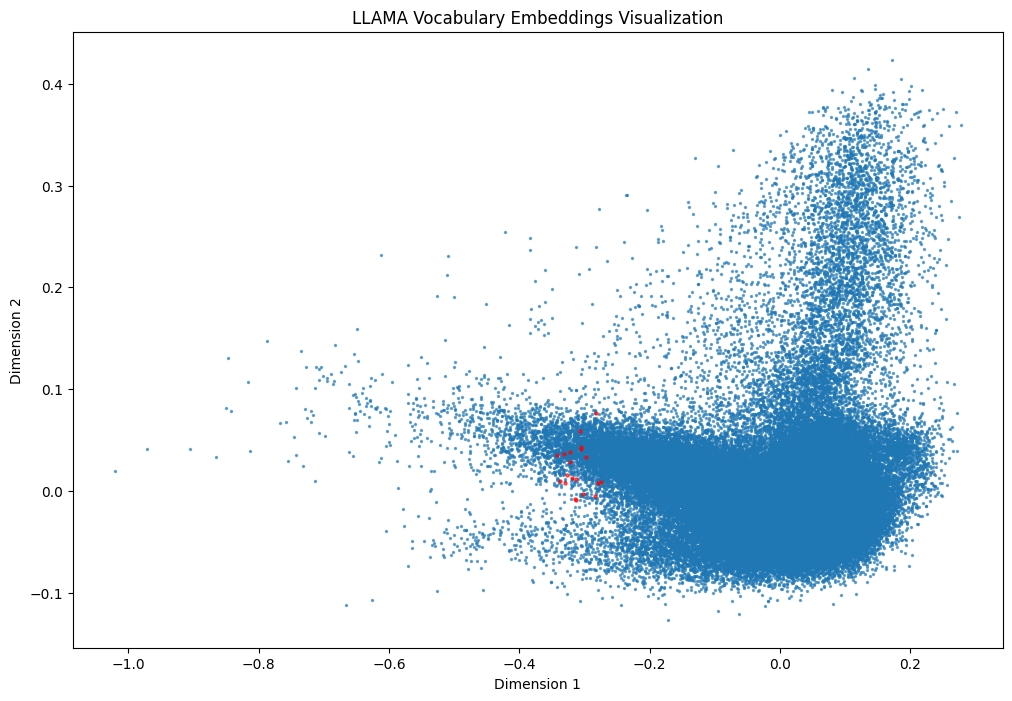}
    \caption{Isotropic Gaussian Prompt Initialization with 5e-4 learning rate (all token embeddings)}
    \label{fig:isotropic-gaussian-5em4}
\end{figure}

\begin{figure}[ht]
    \includegraphics[width=\linewidth]{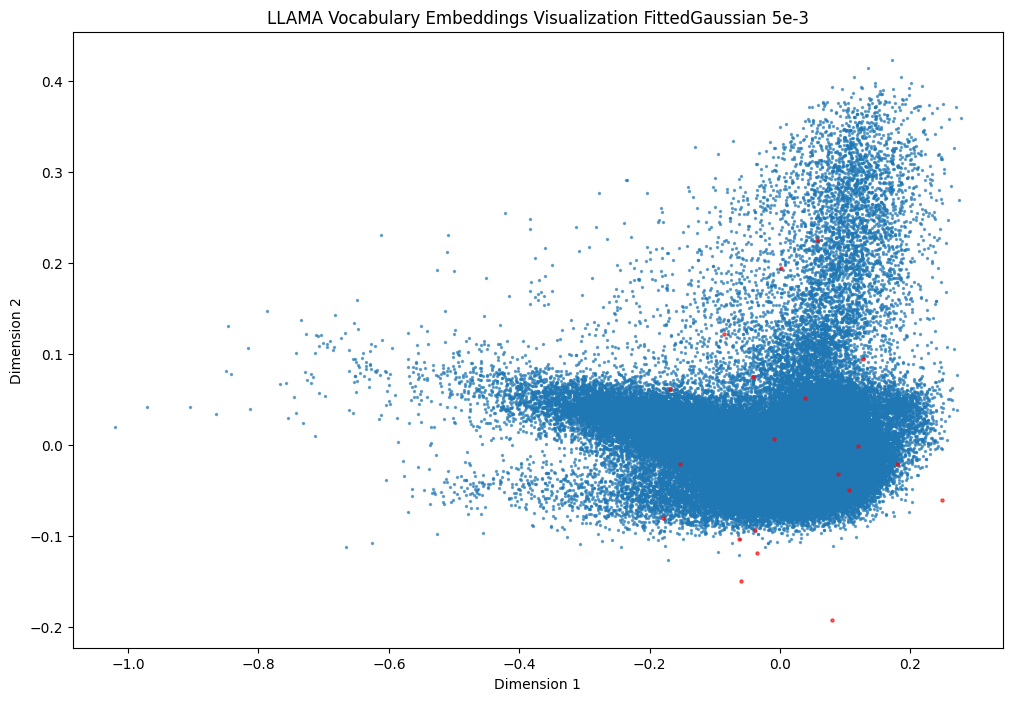}
    \caption{Fitted Gaussian Prompt Initialization with 5e-3 learning rate (all token embeddings)}
    \label{fig:fitted-gaussian}
\end{figure}

To investigate the impact of initialization priors and learning rates on Soft Prompt-Tuning, we conducted a series of three experiments. These experiments employed different configurations for the initialization of prompt embeddings and adjusted learning rates to assess their effects on the training dynamics. Figures~\ref{fig:isotropic-gaussian-5em3}, \ref{fig:isotropic-gaussian-5em4}, and \ref{fig:fitted-gaussian} present two-dimensional visualizations of the embedding spaces, generated through Principal Component Analysis (PCA), to provide a comparative perspective on pre-trained token embeddings and trained prompt embeddings.

In our experiments, the baseline initialization of the prior was implemented using an isotropic Gaussian distribution, specifically \( \mathcal{N}(0, 0.02) \). This initialization assumes that all dimensions of the embedding space are independent and identically distributed. While simple and computationally efficient, such a prior does not capture the inherent structure and correlations present in the pre-trained embedding space. Another baseline initialization regularly used in Deep Prompt-Tuning is default Xavier initialization.

To address this limitation, we compared them with the Gaussian initialization, fitted to the existing distribution of the embeddings. This involved estimating the mean vector and covariance matrix of the pre-trained embeddings. The mean vector encapsulates the central tendency of the embeddings, while the covariance matrix captures the interdependencies and variances across dimensions. By leveraging this fitted Gaussian prior, we aim to achieve a more realistic and representative initialization that aligns better with the structure of the embedding space. This adjustment aims to guide the Prompt-Tuning process more generally, reducing the likelihood of the embedding collapse.

\begin{figure}[ht]
    \includegraphics[width=\linewidth]{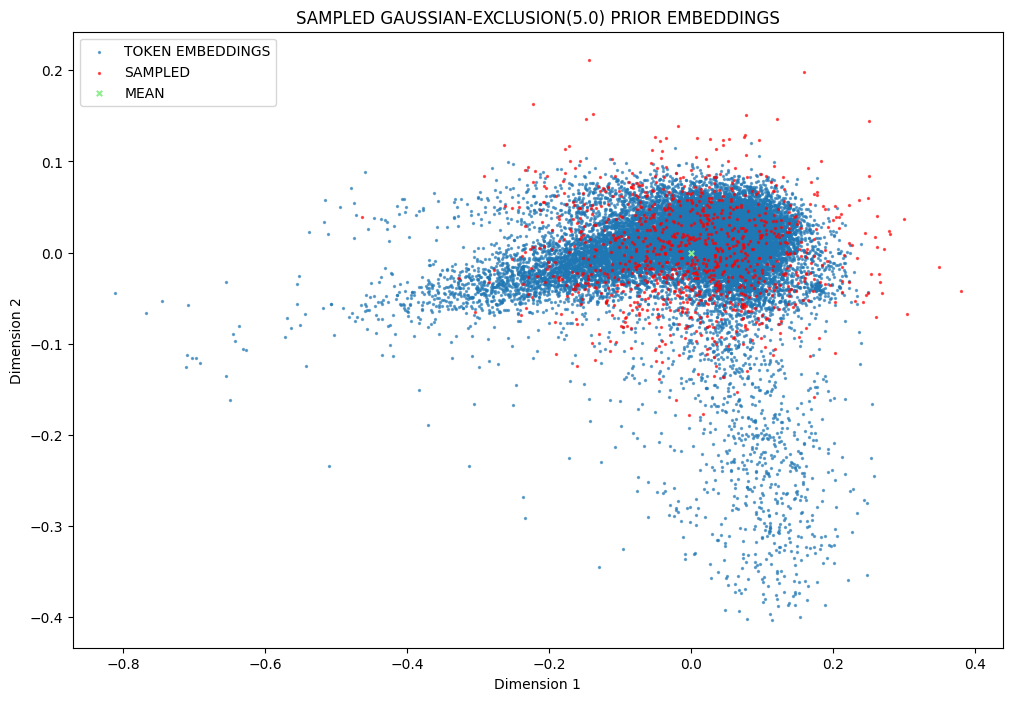}
    \caption{Samples from Gaussian Exclusion Prior. The empirical mean is as a green cross. (all token embeddings)}
    \label{fig:gaussian-exclusion}
\end{figure}

In these figures:

- The red dots represent the trained prompt embeddings, highlighting the outcome of the tuning process.

- The blue dots signify the pre-trained token embeddings, which serve as a reference for the original embedding space.

Contrary to the findings outlined in Section 7 of the Prompt-Tuning paper \cite{liu2021prompt}, where prompt embeddings are often observed to collapse into the existing embedding space, our results suggest a more nuanced behavior. Depending on the initialization strategy (prior) and learning rate, the trained prompt embeddings may exhibit significant divergence from the original embedding space. Certain combinations of priors and learning rates appear to prevent the prompt embeddings from converging into the pre-trained token embedding space, resulting in distinct patterns of divergence.

These findings highlight the sensitivity of Soft Prompt-Tuning to hyperparameter choices, which could have implications for downstream task performance and the interpretability of trained prompts.

\subsection{Gaussian Exclusion Prior}
As in this project we aim to investigate, whether models can work with distant activations, our next approach was to exclude the high density region from sampling distribution with the described Gaussian exclusion prior. An example of such prior is presented in Figure \ref{fig:gaussian-exclusion}. Interestingly, we see how sharply peaked is the token embedding distribution. Training did not lead to any new interesting takeaways, matching the performance. At that point we decided to move on towards optimization on activation spaces.

\subsection{Prior Design Experiments on Activation Spaces}
In this section we present our results as plots on the activation spaces of the models. This means that even for token-level Prompt-Tuning we are plotting the distribution of token-embeddings according to dataset samples, on contrast to plotting the whole token-embedding distribution of the model, where each token appears once. In all experiments we stick to learning rate $\lambda = 0.001$, as a most stable one.
We find that in all our experiments methods match the same final quality of Prompt-Tuning / Deep Prompt-Tuning as in the original initializations. While this is unfortunate in terms of Prompt-Tuning improvements, this result leads us to an interesting observation. In the following sections we are connecting observed trained embeddings distributions with it.  

\begin{figure}[ht]
    \includegraphics[width=\linewidth]{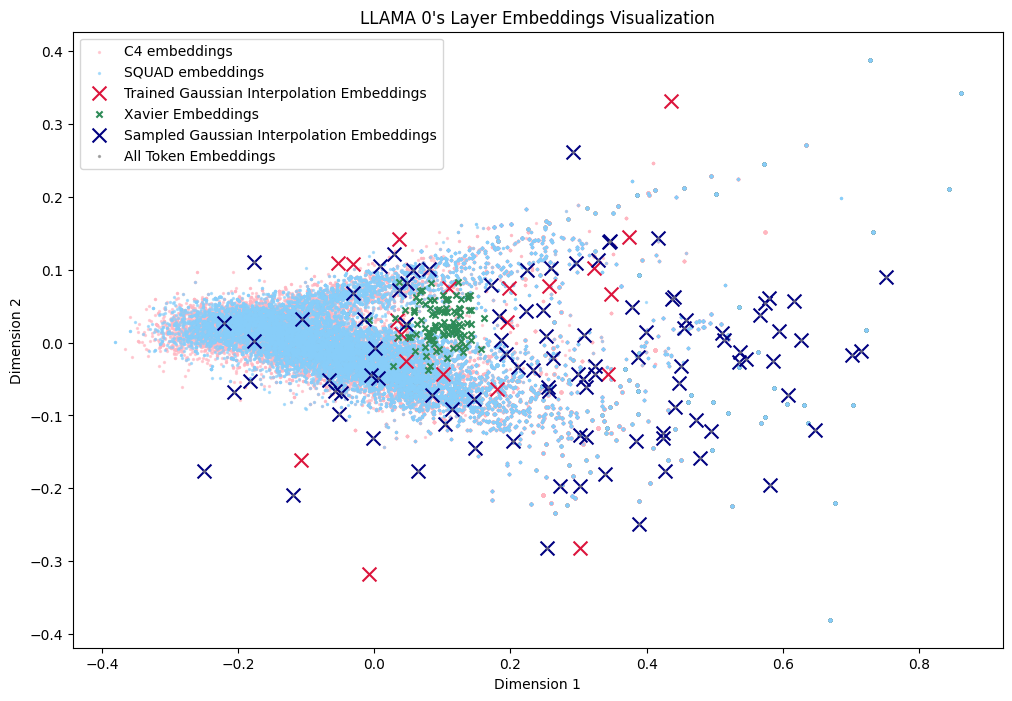}
    \caption{Token-embedding activations on C4 and SQuAD datasets. Sampled and trained embeddings for Gaussian fitted on SQuAD are blue and red correspondingly. Sampled Xavier embeddings are green, trained embeddings lie in the same area.}
    \label{fig:fitted-gaussian-squad}
\end{figure}

Firstly, we present token-level distribution of the model on C4 and SQuAD datasets in Figure \ref{fig:fitted-gaussian-squad}. Along with them, we show the distribution of sampled and trained embeddings using the Gaussian fitted on SQuAD samples. Additionally, we visualize sampled Xavier embeddings as green; trained Xavier embeddings generally do not diverge from the sampled region. 

First conclusion that we could draw from this plot is that Gaussian Fitted initialization does not really represent the activations distribution clearly, which supports our hypothesis on activations distributions being sharply peaked. 

This plot represents an interesting finding that we will continue to observe in further experiments. We can see that some of the trained Gaussian embeddings stayed far away from the activation distribution,  saving the shape of the prior to some degree. Given that the model quality stays the same, we may hypothesize that the model is capable of using Prompt-Tuning embeddings to full extent no matter where they are located particularly. It is important that in Gaussian Fitted and all other "diverged" initializations it takes more time to converge to the same validation loss level, so the search for optimal embeddings in distant regions takes longer. Another important aspect is the presence of several trained embeddings close to the original distribution. We refer to the original Prompt-Tuning paper results that show the insufficiency of very short prompts, so such small intersection is not enough and distant samples contribute to the final performance.  

\begin{figure}[ht]
    \includegraphics[width=\linewidth]{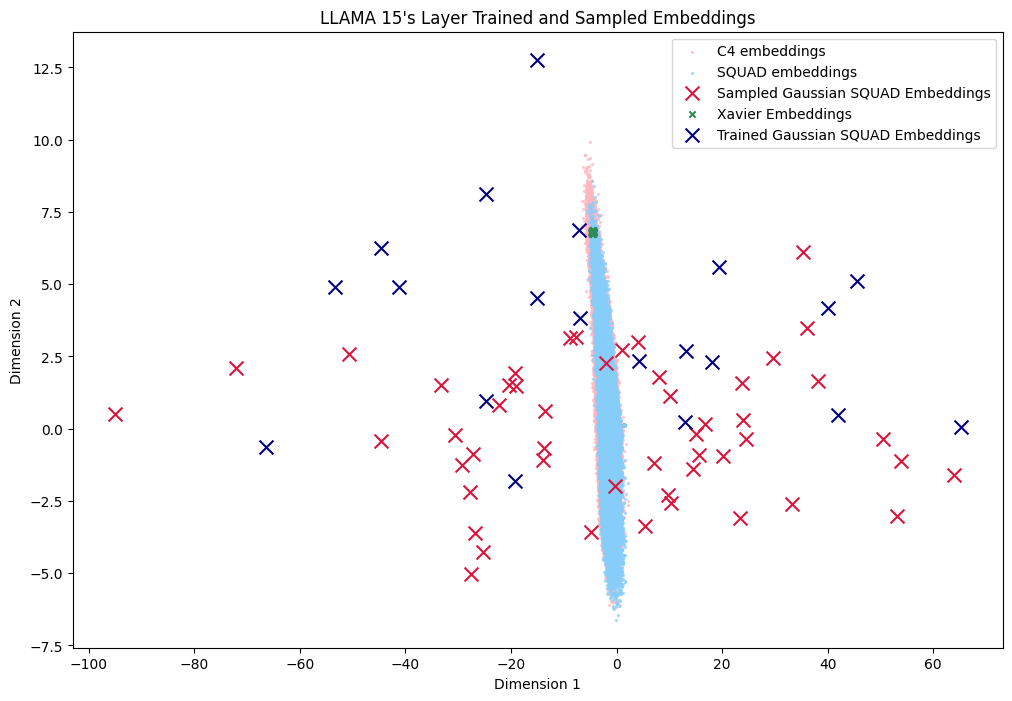}
    \caption{Last layer activations on C4 and SQuAD datasets. Sampled and trained embeddings for Gaussian fitted on SQuAD are blue and red correspondingly. Sampled Xavier embeddings are green, trained embeddings lie in the same area.}
    \label{fig:deep-fitted-gaussian-squad}
\end{figure}

At this point we already observed enough evidence regarding the distant priors, so instead of implementing the previously referenced kNN sampling, we decided to focus on Deep Prompt-Tuning experiments.

Next, we present similar results for last layer activations of the model in Figure \ref{fig:deep-fitted-gaussian-squad}. Xavier init is very narrow (again, its trained embeddings do not diverge far from the sampled ones, which is not included in the figure), while Gaussian is even more diverged. Here we can see clearly that trained embeddings do not intersect initial activations, while matching the validation quality.

\subsection{Experiments on MATH dataset}
\begin{figure}[ht]
    \includegraphics[width=\linewidth]{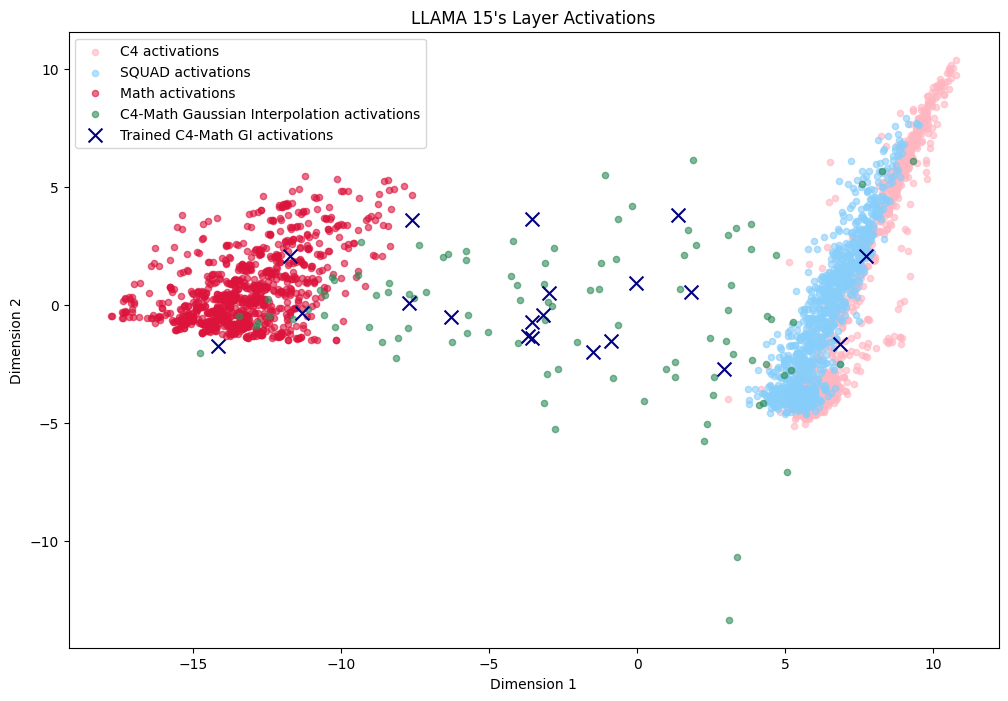}
    \caption{Last layer activations on C4, SQuAD and MATH datasets, together with Interpolated Gaussian prior and posterior}
    \label{fig:deep-math-int}
\end{figure}

We further tried to improve the performance of Prompt-Tuning and control the divergence of embeddings in different task setup. We moved on to experimenting on MATH arithmetics dataset, observing that it has a different cluster of activations, Figure \ref{fig:deep-math-int}. Interestingly, addition of MATH dataset highlights the difference between C4 and SQuAD activations better. 

\begin{figure}[ht]
    \includegraphics[width=\linewidth]{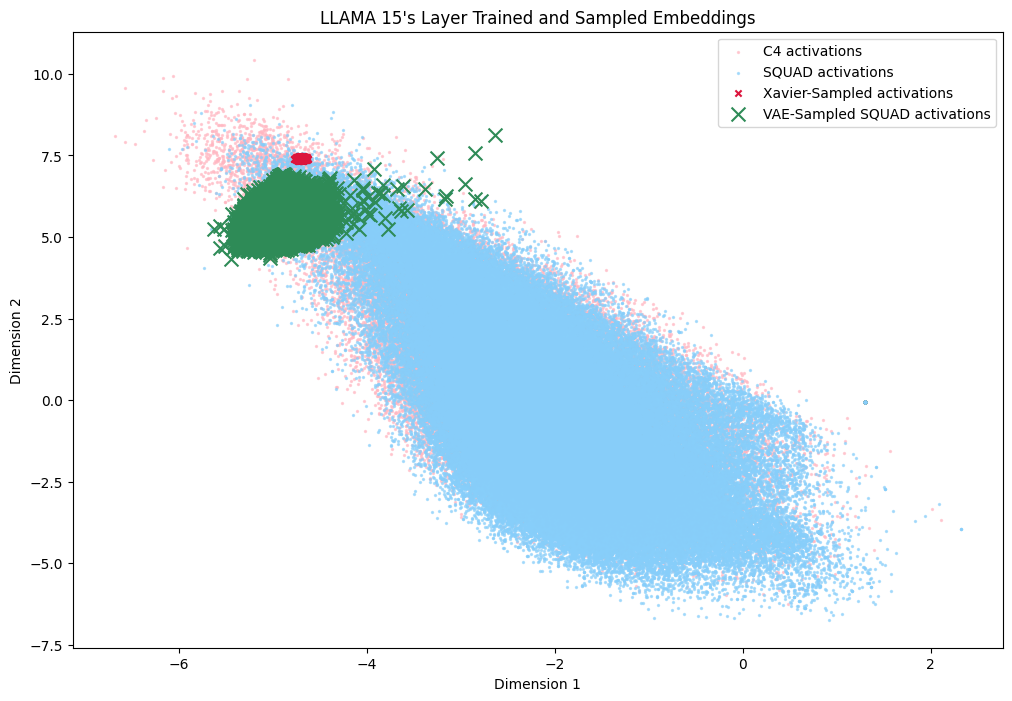}
    \caption{Last layer activations sampled by VAE trained on SQuAD activations}
    \label{fig:deep-vae}
\end{figure}

We interpolate samples between Gaussians fitted on MATH and C4.
The results for this prior and posterior are also presented in Figure \ref{fig:deep-math-int}. The takeaways of this experiment are similar to the previous one. However, this setup is explicitly connected with the stated line of future work on matching activation distributions for different domains. In fact, we provide weak evidence, using Deep Prompt-Tuning, that pre-trained model can find useful intermediate activations between domain clusters. If the results were different, approaching multi-modality from the activation distributions point of view most likely would have been a harder task.

\subsection{VAE Prior}

We tried sampling activations using VAE trained on dataset activations from a pre-trained model, hypothesizing that applying VAE to distinct activation clusters would smoothen the activations distribution. However, despite implementing multiple regularization strategies, the VAE remains prone to collapsing the activation distribution, failing to maintain a structured spread. In the results, SQuAD-trained activations are tightly concentrated in a distinct cluster. Regularizations, such as KL divergence and variance constraints, have not been sufficient to prevent collapse. However, analyzing activation distributions during training process on the level of generative model's latent spaces is an interesting direction of research. Similar topics have been touched in terms of representation learning \cite{vae-repr} and model compression \cite{vae-compression} using VAEs.

\section{Discussion and Further Research}

In this project we have researched the control of Soft and Deep Prompt-Tuning in terms of pre-trained model activations distributions. In this section we discuss our results, looking for connections with different research areas.


\textbf{Embedding Divergence and the Role of Priors}\\
Contrary to the widely observed phenomenon of embedding collapse, our results demonstrate that prompt embeddings do not invariably converge into the pre-trained token embedding space. The same is true for higher-layer activation spaces. Instead, the choice of prior initialization significantly affects whether embeddings diverge or stay within the same cluster. In our experiments we can see that models reach the same validation performance regardless the location of tuned embeddings, which suggests that models are able to use them to full capacity in each of the scenarios. 

Divergence in the embedding space is not necessarily detrimental. On the contrary, it may allow models to better capture task-specific nuances or connect different task domains. However, Prompt-Tuning setup appears to be not enough to connect different clusters. It only serves as a proof of concept, that models could potentially work on activations interpolated between different domains, as we have shown in the Section 4.5. This adaptability aligns with our hypothesis that prompt-tuning posteriors may serve as effective priors for subsequent training tasks.
In future work we may explore the applications of this research in the aforementioned Chains-of-Thought distillation field, which could help design priors for new tokens needed to reduce chain-of-thought length and improve generalizability.


Why do we have a single cluster of activations for fairly different language tasks? The importance of this observation remains unclear. We show that while SQuAD has activations similar to pre-training C4, MATH dataset's activations lie in a distant cluster, and this property holds for activations on all layers. How do generalization abilities of Large Language Models emerge in terms of evolution of their activations during training? Probably, distant clusters on NLP and math tasks bring evidence, that these domains are not yet fully integrated into each other? 

Activation-based point of view is in some sense related to the recently proposed Forward-Forward algorithm \cite{ff}, which proposes to contrastively train neural networks, regularizing activations on each layer for positive samples. While this algorithm does not reach backpropagation performance in NLP domain, the effect of activation regularizations on the speed of generalization abilities emergence remains an interesting topic.

In future work, it would be interesting to build an experimental setup, in which fine-tuned model could learn the same or separate clusters of activations for a new task, depending on the use of regularization methods. We could further compare the integration of a new task to the initial domain of the model in both scenarios. This will however require precise evaluation design, as measurement of domains integration may be a hard task by itself.

\section{Conclusion}
In this project, we explored Prompt-Tuning behavior through the lens of prior and posterior activation distributions. We showed that regardless of the position of sampled embeddings, both Soft and Deep Prompt-Tuning succeed in training them to the same quality. Trained embeddings sampled from different priors do not converge to the pre-trained clusters of activations, highlighting that models could effectively work with the parts of activation spaces that are not covered by the initial data activations. Besides Prompt-Tuning experiments, we showed that generally trajectories that the model generates are not localized in activation space. Instead, activations on distant tasks have distinct clusters (arithmetics and general NLP tasks in our experiments). These observations made us ask questions on the importance of single cluster of activations for generalization abilities of neural networks in the discussion section. 

\bibliographystyle{IEEEtran}
\bibliography{references}

@article{lester2021power,
  author = {Lester, Brian and Al-Rfou, Rami and Constant, Noah},
  title = {The Power of Scale for Parameter-Efficient Prompt Tuning},
  journal = {arXiv preprint arXiv:2104.08691},
  year = {2021}
}

@article{liu2021prompt,
  author = {Liu, Jing and Zhang, Jian and others},
  title = {Prompt-Tuning v2},
  journal = {arXiv preprint arXiv:2110.07602},
  year = {2021}
}

@article{lee2024bayesian,
  author = {Lee, Haeju and Jeong, Minchan and Yun, Se-Young and Kim, Kee-Eung},
  title = {Bayesian Multi-Task Transfer Learning for Soft Prompt Tuning},
  journal = {arXiv preprint arXiv:2402.08594},
  year = {2024}
}

@article{derakhshani2022bayesian,
  author = {Derakhshani, Mohammad Mahdi and Sanchez, Enrique and Bulat, Adrian and da Costa, Victor Guilherme Turrisi and Snoek, Cees G. M. and Tzimiropoulos, Georgios and Martinez, Brais},
  title = {Bayesian Prompt Learning for Image-Language Model Generalization},
  journal = {arXiv preprint arXiv:2210.02390},
  year = {2022}
}

@article{vae-repr,
  author = {Higgins, Irina and Matthey, Loic and Glorot, Xavier and Pal, Arka and Uria, Benigno and Blundell, Charles and Mohamed, Shakir and Lerchner, Alexander},
  title = {Early Visual Concept Learning with Unsupervised Deep Learning},
  journal = {arXiv preprint arXiv:1606.05579},
  year = {2016}
}

@article{vae-compression,
  author = {Louizos, Christos and Ullrich, Karen and Welling, Max},
  title = {Bayesian Compression for Deep Learning},
  journal = {arXiv preprint arXiv:1705.08665},
  year = {2017}
}

@article{ff,
  author = {Hinton, Geoffrey},
  title = {The Forward-Forward Algorithm: Some Preliminary Investigations},
  journal = {arXiv preprint arXiv:2212.13345},
  year = {2022}
}

\end{document}